\documentclass[letterpaper, 10 pt, conference]{ieeeconf}  

\IEEEoverridecommandlockouts                              

\usepackage[T1]{fontenc}
\usepackage{graphicx}
\usepackage{booktabs}
\usepackage{amsmath}
\usepackage{amssymb}
\usepackage{algorithm}
\usepackage{algorithmic}
\usepackage{hyperref}

\title{\LARGE \bf
Spatial Traces: Enhancing VLA Models with \\ Spatial-Temporal Understanding
}

\author{Maxim A. Patratskiy$^{1}$ and 
    Alexey K. Kovalev$^{2, 1}$ and
    Aleksandr I. Panov$^{2, 1}$%
\thanks{$^{1}$MIPT, Dolgoprudny, 141701, Russia $^{2}$AIRI, Moscow, 121170, Russia.}%
}

\begin{document}

\maketitle
\thispagestyle{empty}
\pagestyle{empty}

\begin{abstract}

Vision-Language-Action models have demonstrated remarkable capabilities in predicting agent movements within virtual environments and real-world scenarios based on visual observations and textual instructions. Although recent research has focused on enhancing spatial and temporal understanding independently, this paper presents a novel approach that integrates both aspects through visual prompting.
We introduce a method that projects visual traces of key points from observations onto depth maps, enabling models to capture both spatial and temporal information simultaneously.
The experiments in SimplerEnv show that the mean number of tasks successfully solved increased for 4\% compared to SpatialVLA and 19\% compared to TraceVLA.
Furthermore, we show that this enhancement can be achieved with minimal training data, making it particularly valuable for real-world applications where data collection is challenging. 
The project page is available at \url{https://ampiromax.github.io/ST-VLA}.

\end{abstract}

\section{INTRODUCTION}
The rapid advancement of large language models (LLMs) has expanded their applications across various domains. 
One of the most promising areas of development is robotics and task planning ~\cite{huang2022inner,kovalev2022application,sarkisyan2023evaluation,driess2023palme,grigorev2024common,grigorev2025verifyllm}, where these models are being adapted to generate sequences of manipulator movements in a three-dimensional space based on environmental observations and task specifications~\cite{song2023llm,hu2023look,onishchenko2025lookplangraph,pchelintsev2025lera}. 
This evolution has led to the emergence of Vision-Language-Action (VLA) models~\cite{jiang2022vima,staroverov2023fine,octo_2023,kim24openvla,li2024cogact,qu2025spatialvlaexploringspatialrepresentations}, which demonstrate the ability to predict agent movements in virtual environments through the integration of visual and textual information. 
Furthermore, recent work~\cite{sermanet2024robovqa,bulatov2024mastering,kuratov2024babilong} shows that LLMs can effectively support long-horizon multi-task reasoning, further reinforcing the applicability of such models to embodied decision making.

Current VLA models, such as OpenVLA~\cite{kim24openvla}, have shown promising results in predicting agent movements based on single visual observations and text instructions. 
However, these models often lack comprehensive spatial understanding of the environment.
Recent work, including SpatialVLA~\cite{qu2025spatialvlaexploringspatialrepresentations}, has attempted to address this limitation by incorporating RGB-D images as input.
Experiments have shown that spatial understanding is crucial for successful robot manipulation tasks.

A significant limitation of current approaches is their treatment of manipulation tasks as sequences of independent action predictions. 
Although some methods, such as CogACT~\cite{li2024cogact}, attempt to predict multiple actions simultaneously, they still fail to incorporate the history of agent-environment interactions effectively. 
This limitation becomes particularly apparent in complex manipulation tasks where the temporal context is essential for successful execution.

Recent developments in visual language models (VLMs) have revealed that these models can be effectively prompted with contextual information in visual space. Models may be prompted with predicted traces of the manipulator during task execution~\cite{gu2023rttrajectoryrobotictaskgeneralization}, traces of movements in the environment~\cite{zheng2024tracevlavisualtraceprompting} or even directions of interactions with objects~\cite{nasiriany2024rtaffordanceaffordancesversatileintermediate}. The TraceVLA~\cite{zheng2024tracevlavisualtraceprompting} approach prompts a model with two images: the original and one with applied traces. Although this approach incorporates temporal information through traces, it does not provide the model with any spatial context.

Based on this analysis, we identify a key gap in current methods and motivate our proposed approach.
Existing VLA models focus either on spatial understanding or temporal context, but do not combine both modalities in a unified framework. 
To address these limitations, we propose the \textbf{Spatial Traces}, which introduces a novel visual prompting technique by overlaying manipulator traces on predicted depth maps. The resulting model that uses this technique is ST-VLA.
\textbf{Spatial Traces enables the model to utilize spatial and temporal information simultaneously}, resulting in improved performance in complex manipulation tasks. 
Our empirical results demonstrate significant gains in the SimplerEnv virtual environment over previous methods, particularly in scenarios requiring nuanced spatial-temporal reasoning. The experiments show an increase in 4\% compared to SpatialVLA and 19\% compared to TraceVLA in the number of tasks solved.

Overall, the contributions of our work are as follows:
\begin{enumerate}
    \item We propose a Spatial Traces approach that enables the VLA model to use spatial and temporal information from the environment through a single image.
    \item Experiments in SimplerEnv show that with only 52 training trajectories, our approach improves task success by 4\% over SpatialVLA and 19\% over TraceVLA, demonstrating the advantage of incorporating spatial and temporal information even with minimal fine-tuning.
\end{enumerate}

\section{RELATED WORK}
Recent advances in generalist robotic models have demonstrated notable progress in manipulation tasks. Models such as RT-1-X~\cite{open_x_embodiment_rt_x_2023} and Octo~\cite{octo_2023}, trained on extensive robot datasets, exhibit strong zero-shot generalization across diverse tasks and embodiments. Approaches such as OpenVLA~\cite{kim24openvla} fine-tune vision-language models (LLMs) specifically for robotic manipulation tasks, while CogACT~\cite{li2024cogact} decouples diffusion-based action planning from visual encoding. Despite their robustness, these models typically rely on static observations and do not explicitly incorporate temporal dynamics, restricting their applicability to sequential manipulation scenarios. In contrast, our Spatial Traces approach explicitly encodes temporal information through visual prompting, leveraging historical keypoint trajectories directly embedded into the current observation.

Spatial grounding has become essential for accurate robotic manipulation, with existing methods introducing spatial context through depth maps, voxel grids, or 3D affordances. SpatialVLA~\cite{qu2025spatialvlaexploringspatialrepresentations} integrates depth and egocentric geometry into the model's internal architecture, and SEM~\cite{lin2025sem} leverages multi-view depth maps for comprehensive spatial understanding. 
However, these approaches generally focus on single, static observations, and thus lack the temporal dimension that is crucial for modeling sequential interactions. Our Spatial Traces method addresses this limitation by integrating temporal trajectories with spatial embeddings derived from depth maps, providing simultaneous spatial and temporal grounding.

Visual prompting provides additional contextual guidance for robotic models. RoboTAP~\cite{vecerik2023robotaptrackingarbitrarypoints}, for example, tracks key points via TAPIR~\cite{tapir} to guide agent navigation. TraceVLA~\cite{zheng2024tracevlavisualtraceprompting} overlays visual traces from CoTracker~\cite{cotracker3} on RGB images, enhancing temporal understanding and also providing original observation. RT-Trajectories~\cite{gu2023rttrajectoryrobotictaskgeneralization} predicts manipulator trajectories and visually embeds them into current observations, while RT-Affordance~\cite{nasiriany2024rtaffordanceaffordancesversatileintermediate} visually identifies interaction regions. In parallel, foundation models such as Magma~\cite{yang2025magmafoundationmodelmultimodal} are trained to predict actions and interactions in both digital and physical environments, combining visual and temporal grounding within a unified architecture. Although these methods effectively integrate temporal or affordance information through visual prompts, they lack explicit spatial encoding. Spatial Traces improves these approaches by explicitly embedding key point trajectories onto a depth-informed spatial representation, capturing both historical dynamics and accurate spatial context simultaneously.

In summary, our proposed Spatial Traces approach combines the strengths of existing approaches by integrating both spatial and temporal information through a single visual input. Specifically, we overlay 3D depth-informed embeddings with historical keypoint trajectories tracked across previous observations. This unified spatio-temporal representation significantly enhances generalization to complex, realistic robotic manipulation tasks.

\begin{figure*}[t]   
    \centering
    \includegraphics[width=\linewidth]{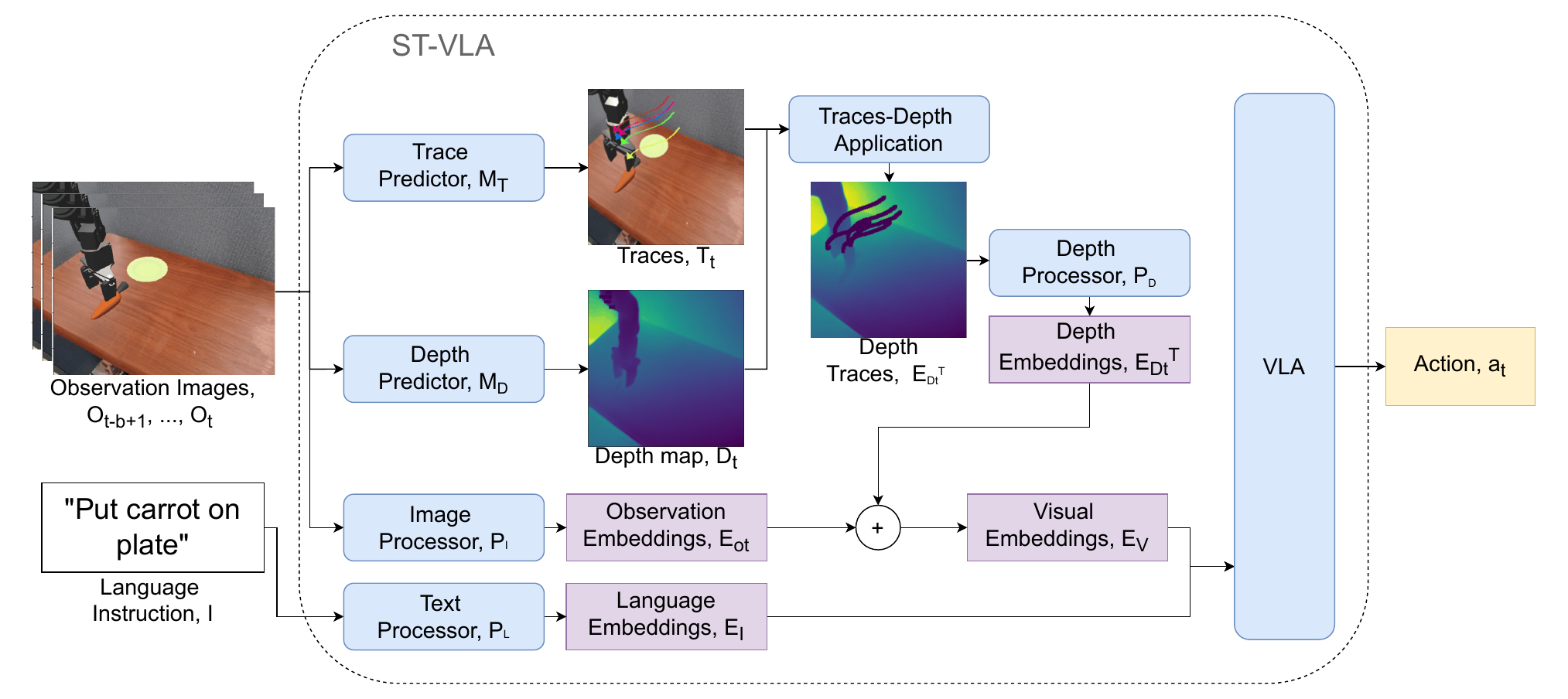}
    \caption{
    The integration of the Spatial Traces mechanism into the VLA model results in the ST-VLA architecture.
    The model receives a textual instruction and a sequence of 30 previous observations as input, and predicts an action in 3D space as a relative displacement from the current position.}
    \label{fig:spatial_trace_vla}
\end{figure*}

\section{METHODOLOGY}
\subsection{Problem Formulation}
Robotic manipulation tasks can be formulated as the objective of reaching a desired environmental state defined by a natural language instruction. Specifically, given an initial state of the environment, the goal is to determine a sequence of actions that an agent performs to transit it and all object states from initial states to goal states.

The manipulation task is defined as a tuple $\mathcal{P} = (S, O, A, T_F, S_I, S_g, I)$,
\noindent where: $S, O, A$ -- are the state, observation and action spaces respectively, $T_F$ -- is a transition function that updates current state with provided action ($s_{t+1} = T(s_t, a_t)$), $S_I, S_G$ -- are initial and goal environment states, $I$ -- text instruction describing goal state. The goal of the manipulation task is to predict with some model $M$ a sequence of actions $[a_0, \dots, a_N]$ based on instruction $I$ that transfer all objects from their initial state to the goal: $[a_0, \dots, a_N] = M(S_I, I)$, $T_F(S_I, [a_0, \dots, a_N]) = S_G$.
The common objective of existing methods is to find a sequence of feasible actions $P = [a_0, \dots, a_T]$ that predict each new action independently using VLA conditioning on current observation and instruction:
\begin{equation}
    \label{eq:vla-step}
    a_t = VLA(o_t, I).
\end{equation}
However, here lies the lack of temporal understanding. The Spatial Traces approach proposes to utilize $b$ previous observations in order to introduce the history of environment-agent interactions into the model input: 
\begin{equation}
    a_t = VLA([o_{t-b+1}, o_{t}], I).
\end{equation}
We aim to optimize the weights of the VLA policy so that it maximizes task success across episodes. Specifically, the training objective is to minimize the mean squared error (MSE) between the predicted model action and the ground truth from the train dataset: $\theta^* = \arg\max_{\theta} \mathbb{E}_{\mathcal{P} \sim \mathcal{D}} [\text{MSE}(\text{VLA}_\theta)]$,
where $\theta$ are the learnable weights of the VLA model and $\mathcal{D}$ denotes the distribution of training tasks. This formulation ensures that the model not only learns to interpret the current observation and instruction, but also integrates relevant temporal context via embedded historical visual information.
\subsection{Spatial Traces}
\label{sec:st-app}
Existing methods utilize spatial information~\cite{qu2025spatialvlaexploringspatialrepresentations}, prompting models with key points or their traces~\cite{nasiriany2024rtaffordanceaffordancesversatileintermediate,gu2023rttrajectoryrobotictaskgeneralization,zheng2024tracevlavisualtraceprompting}, but handle spatial and temporal information separately without integrating them into a unified representation.
The proposed Spatial Traces method integrates two complementary types of visual information to enhance the model's understanding of the environment: \textbf{Depth Information}, \textbf{Visual Traces}. The overall processing pipeline is described in Algorithm~\ref{alg:spatial_traces} and illustrated in Figure~\ref{fig:spatial_trace_vla}.

In our method, we use spatial information collected from depth maps $D$ that can be predicted by any depth estimation model $M_D$. 
To introduce temporal information into the VLA input, we use visual traces $T=[t_1, \dots, t_m]$ of key points of the observations, constructed with model $M_T$. 
Each $t_i$ is a sequence of 2D coordinates of a specific point in the observation images over time $t_i=[(x_0, y_0), \dots,(x_b,y_b)]$, where $b$ denotes the size of the observation buffer $B$. 
Each trace $t_i$ is placed matching their 2D coordinates to the current depth image $D_t$, resulting in $D^T_t$.
The final depth image and original observation are processed with image processors $P_I$ and depth processor $P_D$ to predict embeddings $E_{O_t}$ and $E_{D^T_t}$. 
Then they are summed up, resulting in vision embeddings $E_V$ that are passed to the VLA model with embeddings of text instruction $E_I$. 
According to our method, the implementation Equation~\ref{eq:vla-step} can be reformulated as $a_t = VLA(E_V, E_I)$.

\begin{algorithm}[H]
    \caption{Spatial Traces Algorithm.}
    \label{alg:spatial_traces}
    \begin{algorithmic}[1]
        \REQUIRE Instruction in natural language $I$, buffer size $b$, $b$ previous observations $[o_{t-b+1}, \dots,o_{t}]$
        \ENSURE New action in the environment $a_t$
        \STATE $E_{o_t} := P_I(o_t)$                     \COMMENT{Predicting observation embeddings}
        \STATE $T_t = M_T([o_{t-b+1}, \dots,o_{t}])$     \COMMENT{Predicting key points traces}
        \STATE $D_t := P_D(o_t)$                         \COMMENT{Predicting depth map}
        \STATE $D^T_t = D_t \leftarrow T_t$              \COMMENT{Applying traces to the depth map}
        \STATE $E_{D^T_t} = P_D(D^T_t)$                  \COMMENT{Predicting depth map embeddings}
        \STATE $E_V = E_{O_t} + E_{D^T_t}$               \COMMENT{Summing up visual embeddings}
        \STATE $a_t = \text{VLA}(E_V, E_I)$              \COMMENT{Predicting next step with VLA}
        \RETURN $a_t$
    \end{algorithmic}
\end{algorithm}

\section{IMPLEMENTATION DETAILS}
The action prediction process in our model follows a structured pipeline as described in Section \ref{sec:st-app}. 
To ensure fair comparison with SpatialVLA~\cite{qu2025spatialvlaexploringspatialrepresentations} and TraceVLA~\cite{zheng2024tracevlavisualtraceprompting}, we adopt similar architectural components and input processing procedures in our experiments.
The image processor used $P_I$ is Siglip~\cite{zhai2023sigmoidlosslanguageimage}, the trace predictor $M_T$ is Co-Tracker~\cite{karaev23cotracker}. 
The depth estimation model $M_D$ is ZoeDepth~\cite{https://doi.org/10.48550/arxiv.2302.12288}. 
The following depth processor $P_D$ is the Ego3D Positional Encoder introduced in SpatialVLA.
The action prediction model for robot manipulation is based on the PaliGemma2 vision-language model~\cite{steiner2024paligemma2familyversatile}, initialized from a checkpoint used in the SpatialVLA experiments.
The Co-Tracker model constructs trajectories of key points using the buffered images. It is important to note that Co-Tracker's effectiveness in identifying distinctive points may vary depending on the similarity between consecutive images. In cases where sequential images show minimal differences, which is common in robotic manipulation tasks, Co-Tracker may not initiate trace point construction from the first observations. To mitigate this issue, we do not provide a model with traces for the first five steps in the environment.

\section{EXPERIMENTAL SETUP}
In our experiments, we evaluate the effectiveness of the proposed Spatial Traces method, which integrates historical key point's trajectories into depth maps to enhance spatial and temporal grounding in robotic manipulation. The application of this approach results in the ST-VLA model, which is evaluated in the following experiments.
The goal is to assess how this visual prompting strategy impacts model performance and whether it enables generalization from limited fine-tuning data.

To this end, we conducted a series of experiments designed to answer the following research questions (RQ):
\begin{itemize}
\item \textbf{RQ1:} To what extent does integrating keypoint traces into depth maps improve manipulation performance, and can this improvement be achieved with minimal training data?
\item \textbf{RQ2:} How does the length of the agent's interaction history (i.e., the number of traced steps) affect the quality of model predictions?
\item \textbf{RQ3:} What strategies for embedding trajectories into depth observations are most effective?
\item \textbf{RQ4:} Is the observed performance gain attributable solely to fine-tuning on Bridge data, or is it stem from the trace-based visual prompting?
\end{itemize}

\subsection{Evaluation Environment}
\begin{figure}[t]
    \centering
\includegraphics[width=\linewidth]{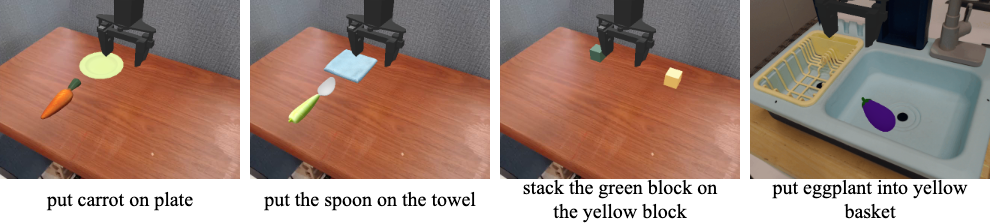}
    \caption{Representative examples of tasks from SimplerEnv. All episodes are based on real robot tasks from the Bridge dataset.}
    \label{fig:simpler_visuals}
\end{figure}

For a quantitative evaluation of the performance of the proposed Spatial Traces approach, we utilized the SimplerEnv~\cite{li24simpler} testing environment, in which the agent is required to follow instructions for manipulating objects within a scene.
This framework, built upon Robosuite~\cite{robosuite2020}, is specifically designed for robot manipulation tasks in virtual environments. 
We selected four representative tasks from the Bridge dataset for evaluation:
\textbf{Put Spoon}, \textbf{Put Carrot}, \textbf{Stack Blocks}, \textbf{Put Eggplant}. 
Examples of visual observations and tasks instructions are shown in Figure~\ref{fig:simpler_visuals}.

\begin{figure}[t]
    \centering
    \includegraphics[width=\linewidth]{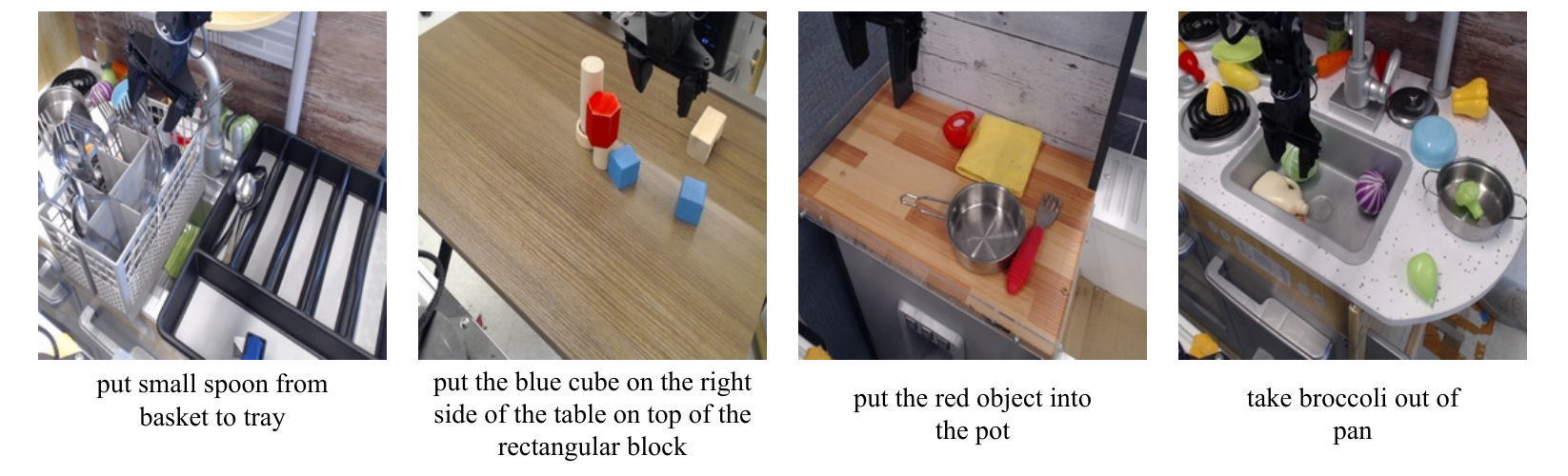}
    \caption{Representative examples of tasks from the Bridge dataset.}
    \label{fig:bridge_data}
\end{figure}

The SimplerEnv framework was chosen for evaluation due to its challenging nature, as demonstrated by the overall low success rates achieved by current methods. 
Previous work~\cite{li2023generalist} has extensively studied the requirements for building effective planners, revealing that modern methods typically achieve success rates below 40\% in Bridge tasks in the SimplerEnv environment.

\subsection{Evaluation Metrics}
We used two primary metrics to assess the quality of task completion: \textbf{Goal Condition Success (GCS)} and \textbf{Success Rate (SR)}. \textbf{GCS} measures the percentage of episodes in which the agent successfully achieved the basic task completion conditions. The condition is a specific state of the environment that can be achieved only after successful execution of several actions. This metric evaluates the model's ability to reach fundamental task objectives, such as grasping and detaching objects from surfaces. For SimplerEnv tasks, all conditions are achieved if the target object was detached from the table.
\textbf{SR} quantifies the percentage of fully completed episodes, considering more stringent criteria that require successful completion of all task stages, achieving the goal state.

\subsection{Train Setup}
\label{sec:train-setup}
In this study, our objective is to demonstrate the effectiveness of enriching visual information and improving task performance with minimal computational resources. 
To enable the model to understand spatial-temporal information, we performed fine-tuning on the Bridge~\cite{walke2023bridgedata} dataset, which contains extensive trajectories of real robot manipulations collected through manual control in various real-world scenarios. 
For our fine-tuning process, we selected a representative subset of 52 trajectories from the original training split. 
Figure~\ref{fig:bridge_data} illustrates examples of visual observations and corresponding text instructions from the dataset.

The training process used 52 trajectories from the Bridge training dataset, resulting in a total of 1,969 steps captured in the real environment. 
To ensure reproducibility and fair comparison, we kept consistent random seeds in all experiments. 
Training examples were presented to all models in the same order. 
We used LoRA~\cite{hu2022lora} adapters applied to all linear layers of the model, with the following hyperparameters: learning rate = 5e-5, batch size = 1, epochs = 2, AdamW~\cite{loshchilov2019decoupledweightdecayregularization} optimizer and LoRA rank = 32. 
The total training time of one model is two hours on a single NVIDIA TITAN RTX.

\subsection{Baselines}
For our experiments, we selected SpatialVLA~\cite{qu2025spatialvlaexploringspatialrepresentations} as the base model due to its strong performance in spatial understanding tasks. The model was pre-trained on 1.1 million trajectories from the Open X-Embodiment~\cite{open_x_embodiment_rt_x_2023} dataset and utilizes PaliGemma2~\cite{steiner2024paligemma2familyversatile} as its visual language model. We maintained the same training and post-processing procedures as described in the original SpatialVLA work to ensure fair comparison.
To evaluate the effectiveness of our approach, we conducted comparative experiments with several popular approaches. 
The following methods were included in our comparison:
OpenVLA~\cite{kim24openvla}, RT-1-X~\cite{open_x_embodiment_rt_x_2023}, Octo~\cite{octo_2023}, RoboVLM~\cite{li2023generalist}.
These baselines were chosen for their strong performance in vision language manipulation tasks and their architectural diversity, providing a comprehensive benchmark for evaluating our method.

\section{EXPERIMENTAL RESULTS}
In this section, we present the results of our experiments designed to evaluate the effectiveness of the proposed Spatial Traces approach in robotic manipulation tasks. The primary focus is to assess whether incorporating keypoint trajectories into depth-based visual inputs can enhance model performance.

\textbf{Q1. To what extent does integrating keypoint traces into depth maps improve manipulation performance, and can this improvement be achieved with minimal training data?} Table~\ref{tab:main-results} presents the comprehensive evaluation results of our proposed method and the baseline approaches on the SimplerEnv. 
The results demonstrate the effectiveness of Spatial Traces approach in various manipulation tasks.
All experiments were conducted in the same manner as in the original work~\cite{qu2025spatialvlaexploringspatialrepresentations}.
Task ``Stack blocks'' is the hardest, with most models achieving 0.0\% SR.
In this scenario, the agent requires a precise position estimation of the placement.
Many models cannot perform such actions, despite grabbing one of the blocks and achieving a high average GCS.

\footnotetext[1]{The baseline models results were taken from SpatialVLA~\cite{qu2025spatialvlaexploringspatialrepresentations}.}
\begin{table}[t]
    \caption{Results of the proposed method and baselines in SimplerEnv.}
    \centering
    \footnotesize
    \label{tab:main-results}
    \resizebox{\linewidth}{!}{%
        \begin{tabular}{lcccccccccc}
            \toprule
            \textbf{Method} & \multicolumn{2}{c}{\textbf{Put spoon}} & \multicolumn{2}{c}{\textbf{Put carrot}} & \multicolumn{2}{c}{\textbf{Stack blocks}} & \multicolumn{2}{c}{\textbf{Put eggplant}} \\
            \cmidrule(lr){2-3} \cmidrule(lr){4-5} \cmidrule(lr){6-7} \cmidrule(lr){8-9}
            & \textbf{GCS$\uparrow$} & \textbf{SR$\uparrow$} & \textbf{GCS$\uparrow$} & \textbf{SR$\uparrow$} & \textbf{GCS$\uparrow$} & \textbf{SR$\uparrow$} & \textbf{GCS$\uparrow$} & \textbf{SR$\uparrow$} \\
            \midrule
            OpenVLA\footnotemark[1] & 4.1 & 0.0 & 33.3 & 0.0 & 12.5 & 0.0 & 8.3 & 4.1 \\
            RT-1-X\footnotemark[1] & 16.7 & 0.0 & 20.8 & 4.2 & 8.3 & 0.0 & 0.0 & 0.0 \\
            Octo\footnotemark[1] & 34.7 & 12.5 & 52.8 & 8.3 & 31.9 & 0.0 & 66.7 & 43.1 \\
            Octo-Small\footnotemark[1] & 77.8 & 47.2 & 27.8 & 9.7 & 40.3 & 4.2 & 87.5 & 56.9 \\
            RoboVLM\footnotemark[1] & 37.5 & 20.8 & 33.3 & 25.0 & 8.3 & 8.3 & 0.0 & 0.0 \\
            \midrule
            SpatialVLA             & 18.2          & 18.2          & 45.5          & \textbf{36.4} & 81.8          & \textbf{45.5} & 54.5          & 36.4          \\
            TraceVLA               & 27.3          & 9.1          & 36.4          & 9.1          & 81.8          & 0.0          & \textbf{90.9} & \textbf{72.7}          \\
            ST-VLA                 & \textbf{36.4} & \textbf{36.4} & \textbf{50.0} & 25.0          & \textbf{90.9} & 36.4          & 81.8          & 54.5          \\
            \bottomrule
        \end{tabular}%
    }
\end{table}

\begin{figure}[t]
    \centering
    \includegraphics[width=0.8\linewidth]{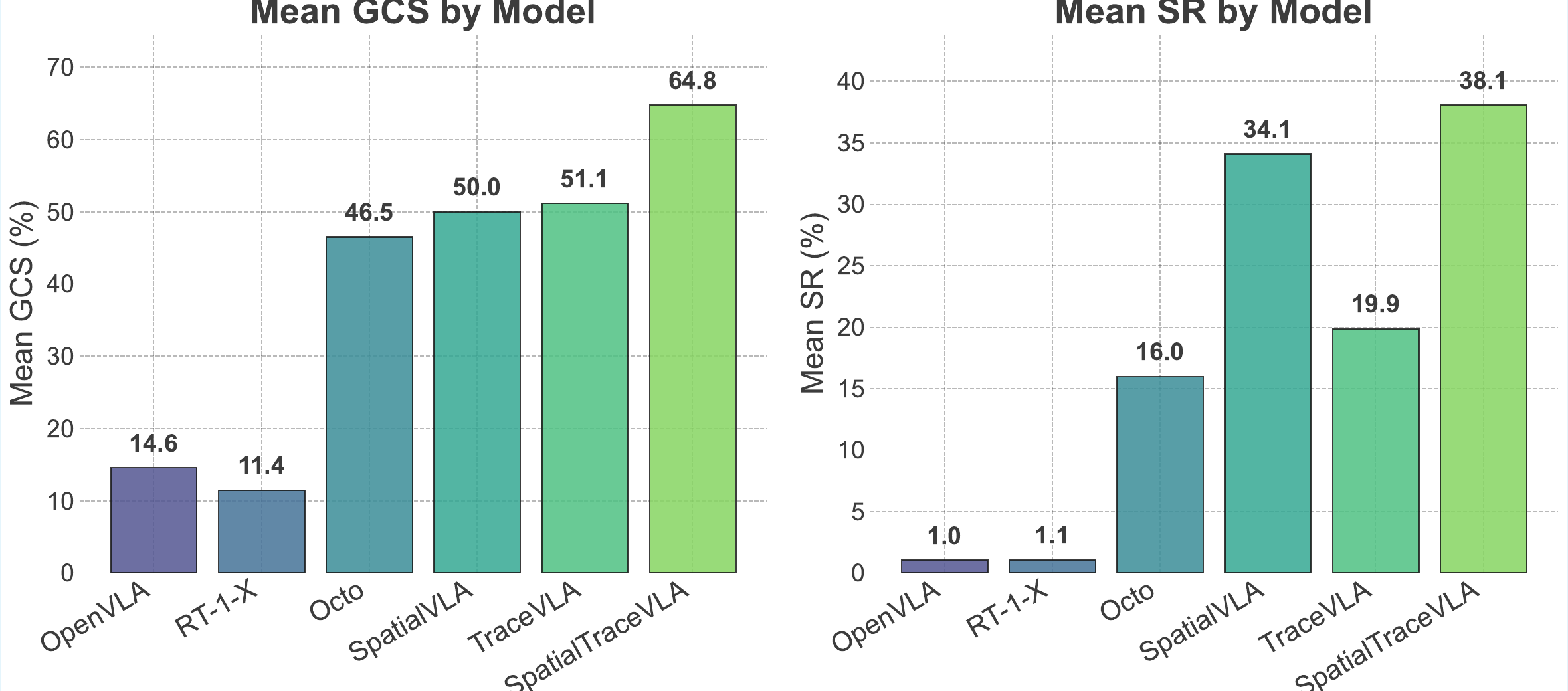}
    \caption{Average GCS and SR achieved by each method.}
    \label{fig:main_results_plot}
\end{figure}

Figure \ref{fig:main_results_plot} presents the average GCS and SR achieved by each method. 
The Spatial-Trace model shows a notable improvement in GCS, outperforming SpatialVLA by 14.8\% and TraceVLA by 13.7\%. 
Furthermore, the SR metric increases by 4\% compared to SpatialVLA and by 18.7\% compared to TraceVLA.

\begin{figure}[t]
    \centering
    \includegraphics[width=1\linewidth]{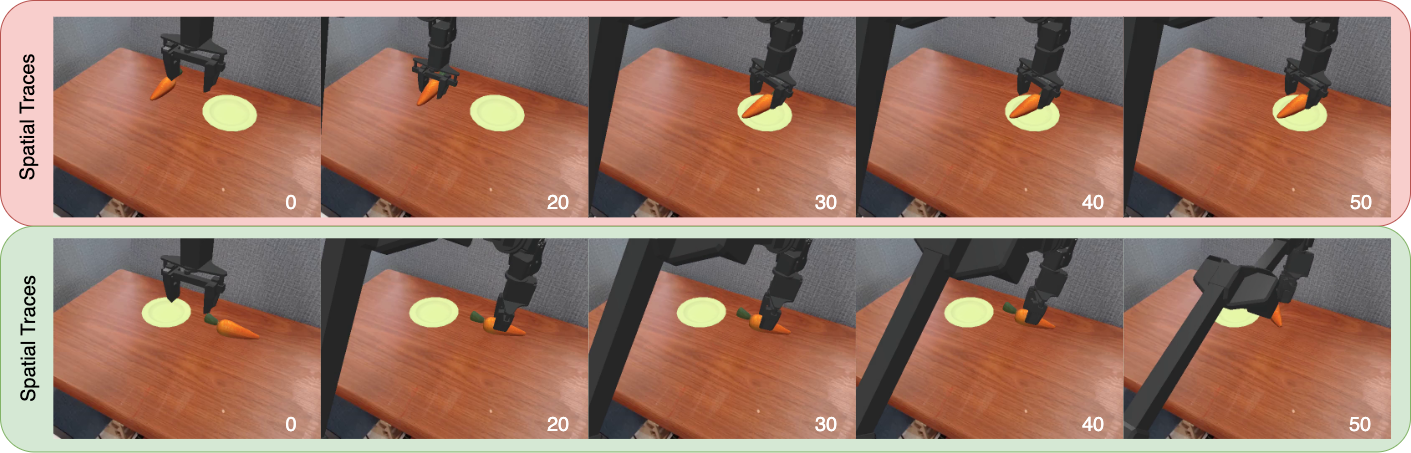}
    \caption{Example of successful and failed execution of ``put carrot on plate'' task by ST-VLA. First row represents completed and second row failed sequences. Numbers on images present the index of step, where observation was taken from. In the first trial, the agent successfully placed the carrot on the plate. In the second, it became stuck during the gripper rotation, resulting in less informative traces and causing a failure in task execution.
}
    \label{fig:failure-analysis}
\end{figure}

Figure \ref{fig:failure-analysis} presents two episodes of successful and failed execution of the ``put carrot on plate'' tasks. Tasks differ by initial object states. The main dissimilarity in the carrot position is in the first run, where it is parallel to the gripper, and in the second orthogonal. The first episode requires only transition, despite the second one with additional rotation. The likely reason for this failure is slow rotation, where consecutive observations differ very little from one another. As a result, the extracted traces are either very short or entirely absent. This limits the effectiveness of the Spatial Traces method, as it provides little to no additional information in such scenarios.

\begin{figure}[ht!]
    \centering
    \includegraphics[width=1\linewidth]{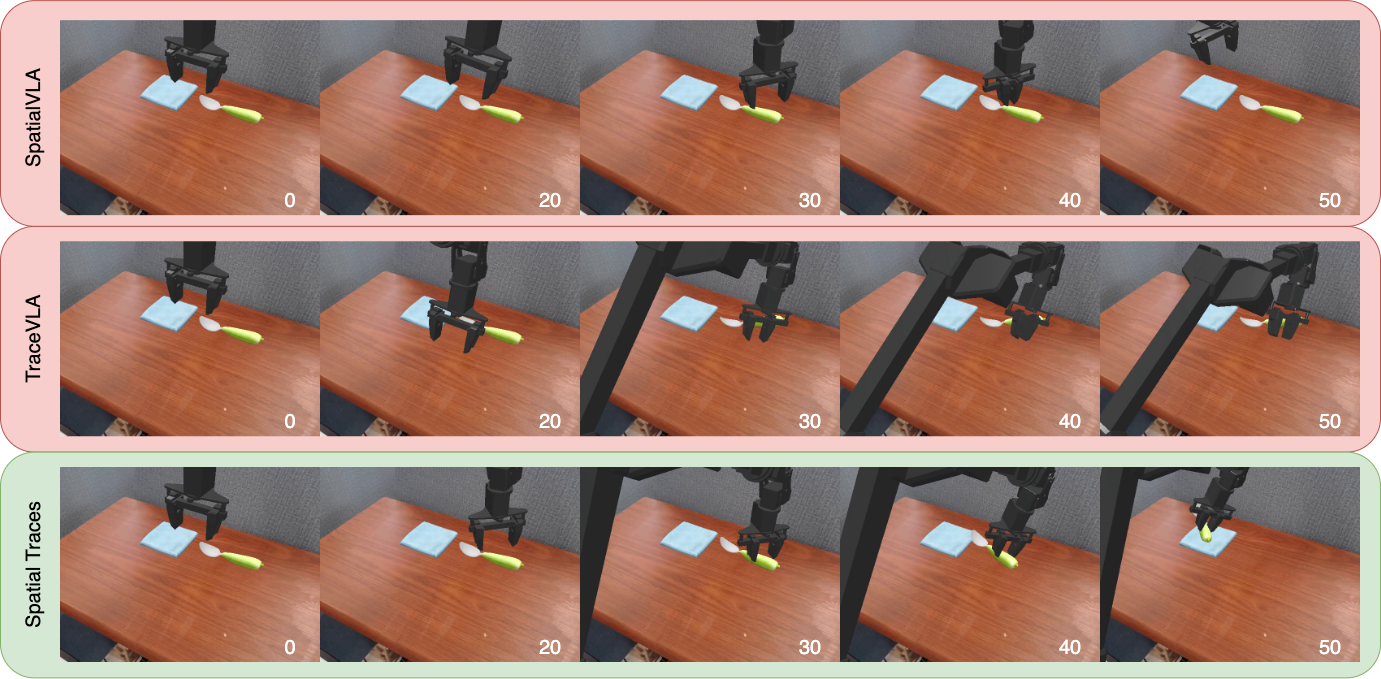}
    \caption{Comparison of SpatialVLA, TraceVLA and ST-VLA executions on ``Put spoon''. Numbers on images present the index of step, where observation was taken from. Baseline models cannot predict spoon location. ST-VLA successfully picks and places the spoon on the towel.}
    \label{fig:main-res-comparison}
\end{figure}

However, some tasks with gripper rotation were completed successfully and better than SpatialVLA and TraceVLA.
Figure~\ref{fig:main-res-comparison} depicts the execution of the same ``put spoon on towel'' task. The first row, is the SpatialVLA, the second is the TraceVLA, and the last one is the ST-VLA. 
The presented approach solves the task, despite being stuck in the first half. After the fast movement of the spoon was performed, the model clearly understood the situation, grabbed the spoon, and successfully placed it on the towel. It should be mentioned that the ST-VLA was trained only on 52 episodes of manipulation tasks. This proofs that spatial-temporal information in depth maps with traces can be introduced to the model with a tiny amount of fine-tuning data.

\textbf{Q2. How does the length of the agent's interaction history (i.e., the number of traced steps) affect the quality of model predictions?} To evaluate the impact of temporal information on model performance, we conducted experiments with the size of the interaction history. 
Table~\ref{tab:ablation-results} presents the results of these experiments. 
These parameters determine the size of the context used to track the points and the applied length of the traces.

\begin{table}[t]
    \caption{Influence of temporal information size on model performance.}
    \centering
    \footnotesize
    \label{tab:ablation-results}
    \resizebox{\linewidth}{!}{%
        \begin{tabular}{lcccccccccc}
            \toprule
            \textbf{Buffer size} & \multicolumn{2}{c}{\textbf{Put spoon}} & \multicolumn{2}{c}{\textbf{Put carrot}} & \multicolumn{2}{c}{\textbf{Stack blocks}} & \multicolumn{2}{c}{\textbf{Put eggplant}} \\
            \cmidrule(lr){2-3} \cmidrule(lr){4-5} \cmidrule(lr){6-7} \cmidrule(lr){8-9}
            & \textbf{GCS$\uparrow$} & \textbf{SR$\uparrow$} & \textbf{GCS$\uparrow$} & \textbf{SR$\uparrow$} & \textbf{GCS$\uparrow$} & \textbf{SR$\uparrow$} & \textbf{GCS$\uparrow$} & \textbf{SR$\uparrow$} \\
            \midrule
            7 images   & \textbf{54.5} & 27.3          & 36.4          & 18.2          & 72.7          & 27.3          & 54.5          & \textbf{54.5} \\
            15 images  & 27.3          & 18.2          & 45.5          & \textbf{36.4} & 72.7          & 9.1           & 63.6          & 36.4          \\
            30 images  & 36.4          & \textbf{36.4} & \textbf{50.0} & 25.0          & \textbf{90.0} & \textbf{36.4} & \textbf{81.8} & \textbf{54.5} \\
            \bottomrule
        \end{tabular}%
    }
\end{table}

The results presented in Table~\ref{tab:ablation-results} indicate a clear relationship between the size of the temporal buffer and the performance of the model. 
In particular, while a short interaction history (buffer size 7) can yield competitive results in certain cases (e.g., high GCS in ``Put spoon''), it generally fails to provide consistent performance across tasks. 
In contrast, longer histories (buffer size 30) lead to more stable results, especially in tasks requiring high spatial understanding, such as ``Stack blocks'' and ``Put eggplant'', where both GCS and SR reach their highest values.
These findings suggest that temporal information accumulated over a longer interaction window enhances the model's ability to reason about dynamic scene changes and improves action planning accuracy. 

\begin{figure}[t]
    \centering
    \includegraphics[width=1.0\linewidth]{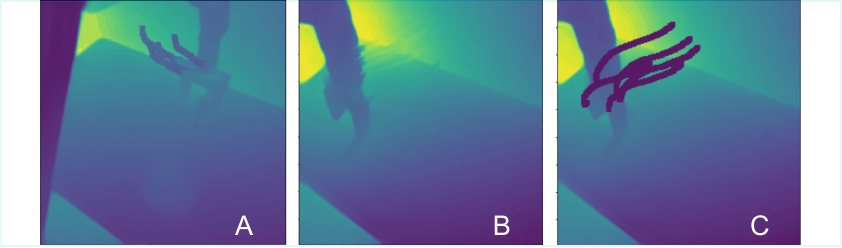}
    \caption{Comparison of trace rendering strategies: (A) depth from the trace’s own observation, (B) depth from the previous observation, (C) depth of the nearest object in the current frame.}
    \label{fig:trace-variants}
\end{figure}

\textbf{Q3. What strategies for embedding trajectories into depth observations are most effective?} The method of visual prompting significantly influences model performance in all tasks. 
We conducted experiments with two distinct approaches to applying traces to the original observation, as illustrated in Figure \ref{fig:trace-variants}. In experiments, variants A and C were used.

We compare different variants of how trace information is incorporated into
the input, relative to the original observation. In the variant A, each trace pixel is
assigned the same depth value as in the corresponding observation, effectively reflecting
the manipulator’s depth at that moment. In the variant B, traces are rendered
using the depth from the preceding observation, which often results in values close to
the background depth. In the variant C, each trace pixel is assigned the depth of
the nearest object present in the current depth map. Due to the fact that variant B was not informative it was excluded from the testing, variants A and C showed promising results, as detailed in Table \ref{tab:trace-variants-results}.

\begin{table}[t]
    \caption{Impact of application methods on model performance for various buffer sizes.}
    \centering
    \footnotesize
    \label{tab:trace-variants-results}
    \resizebox{\linewidth}{!}{%
        \begin{tabular}{lcccccccccc}
            \toprule
            \textbf{ST-VLA} & \multicolumn{2}{c}{\textbf{Put spoon}} & \multicolumn{2}{c}{\textbf{Put carrot}} & \multicolumn{2}{c}{\textbf{Stack blocks}} & \multicolumn{2}{c}{\textbf{Put eggplant}} \\
            \cmidrule(lr){2-3} \cmidrule(lr){4-5} \cmidrule(lr){6-7} \cmidrule(lr){8-9}
            & \textbf{GCS$\uparrow$} & \textbf{SR$\uparrow$} & \textbf{GCS$\uparrow$} & \textbf{SR$\uparrow$} & \textbf{GCS$\uparrow$} & \textbf{SR$\uparrow$} & \textbf{GCS$\uparrow$} & \textbf{SR$\uparrow$} \\
            \midrule
            Type A, buffer 15  & \textbf{36.4}   & 9.1           & 45.5          & 18.2          & \textbf{90.9} & \textbf{36.4} & 63.6          & 45.5          \\
            Type A, buffer 30  & \textbf{36.4}   & 9.1           & 45.5          & \textbf{36.4} & 63.6          & 18.2          & 54.5          & \textbf{54.5} \\
            Type C, buffer 15  & 27.3            & 18.2          & 45.5          & \textbf{36.4} & 72.7          & 9.1           & 63.6          & 36.4          \\
            Type C, buffer 30  & \textbf{36.4}   & \textbf{36.4} & \textbf{50.0} & 25.0          & \textbf{90.9} & \textbf{36.4} & \textbf{81.8} & \textbf{54.5} \\
            \bottomrule
        \end{tabular}%
    }
\end{table}

\begin{figure}[b]
    \centering \includegraphics[width=1.0\linewidth]{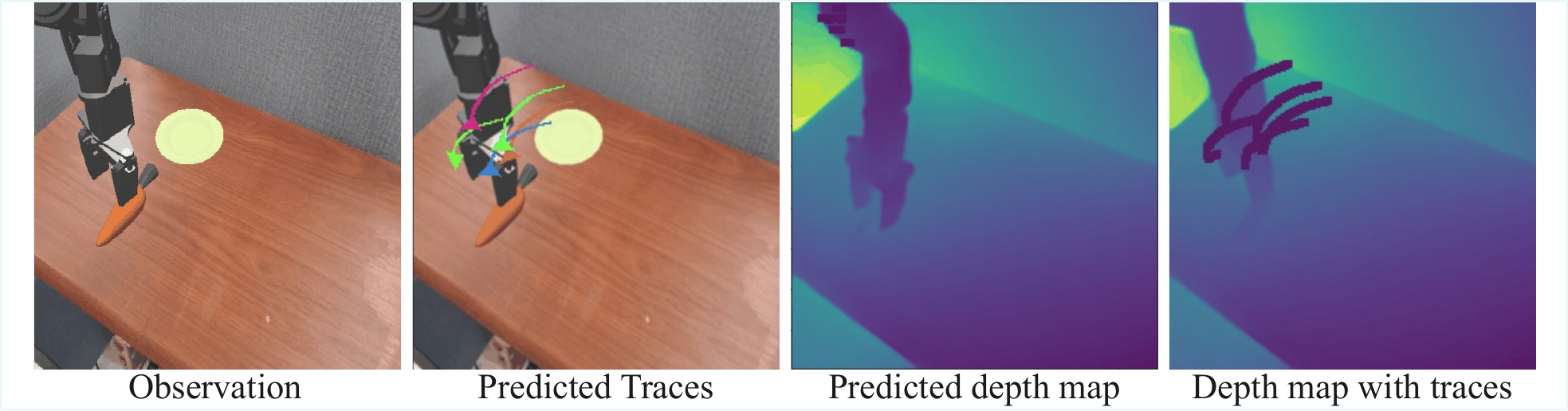}
    \caption{Visualization of depth map processing in the Spatial-Traces approach. This image illustrates the appearance of a raw depth map and the corresponding traces in the original observation. The arrows on the predicted traces indicate the direction of movement. Note, that the traces incorporated into the depth maps do not include directional arrows.}
    \label{fig:simpler-obs-results}
\end{figure}

The results clearly demonstrate the superiority of Variant C over Variant A. This can be attributed to the fact that in Variant A, the model may interpret traces as parts of the manipulator, making it difficult to separate spatial and temporal information. In contrast, Variant C's traces have equal values, are more distinguishable from the background, and can more effectively capture the model's attention. Figure~\ref{fig:simpler-obs-results} provides an example of how traces may look in the original observation and how they are applied to the depth map.

\textbf{Q4. Is the observed performance gain attributable solely to fine-tuning on Bridge data, or does it stem from the trace-based visual prompting?} To address potential concerns regarding the impact of fine-tuning on the Bridge dataset, we conducted additional experiments to evaluate the effect of fine-tuning on model performance.
\begin{table}[t!]
    \caption{Fine-tuning impact on model performance.}
    \centering
    \footnotesize
    \label{tab:finetuning-results}
    \resizebox{\linewidth}{!}{%
        \begin{tabular}{lcccccccccc}
            \toprule
            \textbf{Model} & \multicolumn{2}{c}{\textbf{Put spoon}} & \multicolumn{2}{c}{\textbf{Put carrot}} & \multicolumn{2}{c}{\textbf{Stack blocks}} & \multicolumn{2}{c}{\textbf{Put eggplant}} \\
            \cmidrule(lr){2-3} \cmidrule(lr){4-5} \cmidrule(lr){6-7} \cmidrule(lr){8-9}
            & \textbf{GCS$\uparrow$} & \textbf{SR$\uparrow$} & \textbf{GCS$\uparrow$} & \textbf{SR$\uparrow$} & \textbf{GCS$\uparrow$} & \textbf{SR$\uparrow$} & \textbf{GCS$\uparrow$} & \textbf{SR$\uparrow$} \\
            \midrule
            Base model     & 18.2           & 18.2          & 45.5          & \textbf{36.4} & 81.8          & \textbf{45.5} & 54.5          & 36.4          \\
            Finetuned      & 9.1            & 0.0           & 36.4          & \textbf{36.4} & 63.6          & 18.2          & 72.7          & 36.4          \\
            Traces 0-shot  & 9.1            & 0.0           & 27.3          & 18.2          & 81.8          & \textbf{45.5} & 54.5          & 45.5          \\
            ST-VLA         & \textbf{36.3}  & \textbf{36.4} & \textbf{50.0} & 25.0          & \textbf{90.9} & 36.4          & \textbf{81.8} & \textbf{54.5} \\
            \bottomrule
        \end{tabular}%
    }
\end{table}
The results in Table~\ref{tab:finetuning-results} demonstrate that direct fine-tuning on the Bridge sample leads to a significant decrease in performance, with a mean SR decrease of 10\%. In contrast, SpatialVLA with traces in zero-shot mode shows better performance, with a mean SR decrease of only 6.8\% and improved performance on the ``Put Eggplant'' task. These findings suggest that the enhanced performance of ST-VLA is not primarily due to fine-tuning on the Bridge dataset but rather stems from its ability to effectively utilize temporal information from traces.

\section{CONCLUSION}
In this paper, we introduce the Spatial Traces method, enhancing Vision-Language-Action models by integrating spatial and temporal understanding through visual prompting techniques.
Our experiments demonstrate that the Spatial Traces approach substantially improves performance in manipulation tasks, yielding a 4\% increase compared to SpatialVLA and a 19\% improvement over TraceVLA.
Key findings include the successful extension of visual prompting techniques to depth maps, notable performance gains from fine-tuning on minimal training data, and identification of optimal trace application methods that effectively balance spatial and temporal information.
However, our approach faces limitations related to the dependency on accurate depth map predictions and the effectiveness of trace generation in identifying distinctive points. 
Traces may also occlude target objects, complicating visibility and distance estimation, and fail to provide sufficient interactional cues when the agent moves directly toward the camera.
Addressing these limitations constitutes our future work. 
We aim to explore integration with 3D reconstructed observations, richer temporal visual prompts, adaptive trace generation methods, and application to more complex manipulation scenarios and real-world robotics tasks.

\bibliographystyle{IEEEtran}
\bibliography{IEEEexample}

\end{document}